\documentclass{article}
\usepackage{natbib}
\usepackage{hyperref}
\usepackage{amsmath}
\usepackage{multirow}
\usepackage{url}
\usepackage{epsfig}
\usepackage{graphicx}
\usepackage{subfigure}
\usepackage{algorithm,algorithmic}
\newcommand{\eat}[1]{}

\usepackage{icml2014} 

\input{amssymb.mac}

\icmltitlerunning{A Quantitative Evaluation Framework for Missing Value Imputation Algorithms}

\begin{document} 

\twocolumn[
\icmltitle{A Quantitative Evaluation Framework for Missing Value Imputation Algorithms}
{\bf \icmlfullauthorlist{}}
\vskip 0.3in
]

\begin{abstract}
We consider the problem of quantitatively evaluating missing value imputation algorithms. Given a dataset with missing values and a choice of several imputation algorithms to fill them in, there is currently no principled way to rank the algorithms using a quantitative metric. 
We develop a framework based on treating imputation evaluation as a problem of comparing two distributions and show how it can be used to compute quantitative metrics. We present an efficient procedure for applying this framework to practical datasets, demonstrate several metrics derived from the existing literature on comparing distributions, and propose a new metric called Neighborhood-based Dissimilarity Score which is fast to compute and provides similar results. Results are shown on several datasets, metrics, and imputations algorithms. 
\end{abstract}

\section{Introduction}
\label{sec:intro}


Many commonly used machine learning algorithms assume that the input data matrix does not have any missing entries. In practice this assumption is often violated. \emph{Imputation} is one popular approach to handling missing values -- it fills in the dataset by inferring them from the observed values. Once filled, the dataset can then be passed on to an ML algorithm as if it were fully observed. \emph{Multiple imputation} \cite{lit86} takes into account the uncertainty in the missing values by sampling each missing variable from an appropriate distribution to produce multiple filled-in versions of the dataset. Multiple imputation has been an active area of research in Statistics and ML \cite{bur10,buu99,gel01,li12,rag01,su11,temp09,tem11,buu07,buu11} and has been implemented in several popular Statistics/ML packages \cite{su11,buu11}.

Although there has been much work on multiple imputation, there is little work on the quantitative measurement of the relative performance of a set of imputation algorithms on a given dataset. In the special case where it is known ahead of time that the dataset is to be used only for a specific task (e.g. classification), one can use the evaluation metric for that task (e.g. 0/1 loss) to indirectly measure the performance of an imputation algorithm. But here we are interested in directly evaluating the quality of the imputation, without using a subsequent task for a proxy evaluation.


Current procedures available for evaluating imputations are qualitative. Some imputation tools, such as the R packages \emph{mi} and \emph{MICE}, allow visual comparison of the histogram of observed values and imputed values for each column of the dataset. Such comparisons can be done only for one or two variables at a time and so is at best a weak sanity check. van Buuren and Groothuis-Oudshoorn mention criteria for checking whether the imputations are plausible, e.g. \emph{imputations should be values that could have been obtained had they not been missing}, \emph{imputations should be close to the data}, and they should \emph{reflect the appropriate amount of uncertainty about their `true' values}. But such criteria have not yet been translated into quantitative metrics.


One possibility for quantifying imputation performance is to hold out a subset of observed values from the dataset, impute them, and measure the ``reconstruction error'' between the known and predicted values using an appropriate distance metric, e.g. Euclidean distance for real-valued variables and Hamming distance for categorical variables. As Rubin explains \cite{rub96}, this is not a good metric because there can be uncertainty in the value of the missing variables (conditioned on the values of the observed variables), and directly comparing imputed samples to a single groundtruth value cannot measure this uncertainty.


The key quantity needed to evaluate an imputation is the \emph{distribution of the missing variables conditioned on the observed variables}.  Consider a data vector $Y$ with observed variables $Y_{obs}$ and missing variables $Y_{mis}$, with $Y = \{Y_{obs},Y_{mis}\}$. Let $R$ be a vector of indicator variables indicating which variables in $Y$ are missing or observed. As explained in Little and Rubin \cite{lit86} (Chapter 11, page 219, equation 11.4), the joint distribution of $Y$ and $R$ can be written using Bayes rule, as
\begin{equation}
P(Y,R) = P(Y_{mis}|Y_{obs},R)P(Y_{obs},R).
\end{equation}
A dataset with missing values contains samples for $Y_{obs}$ and $R$ generated according to some ground truth joint distribution $P_{true}(Y,R)$. An imputation algorithm generates samples from an estimate of $\hat{P}(Y_{mis}|Y_{obs},R)$. Note $P(Y_{obs},R)$ does not depend on imputation. Thus, if we can compare $P_{true}(Y_{mis}|Y_{obs},R)$ to $\hat{P}(Y_{mis}|Y_{obs},R)$ that is estimated by an imputation algorithm, then it is possible to quantify how good the imputation is. Therefore the imputation evaluation problem can be turned into a problem of comparing two conditional distributions. In this work we use this connection between imputation evaluation and distribution comparison to develop a quantitative framework for evaluating imputations that can be tractably applied to real-world datasets.

\noindent\textbf{Problem setup:} We have 1) a dataset with values missing according to some (possibly unknown) missingness type, and 2) the output of $K$ different imputation algorithms, with each one producing multiple filled-in versions of the dataset. The goal is to rank the $K$ algorithms according to a chosen metric. Several metric choices are discussed in section~\ref{sec:evaluation}.

\noindent\textbf{Assumption:} It should be possible to estimate a model of $P_{true}(Y_{mis}|Y_{obs},R)$ from the data for each unique missingness pattern $R$. Section~\ref{sec:evaluation} shows how this can be done tractably. The samples given by an imputation algorithm from its estimated conditional $\hat{P}(Y_{mis}|Y_{obs},R)$ are then compared against this model, rather than the original data itself.

\noindent\textbf{Algorithm:} The main steps are the following:
\begin{itemize}
\item A) Construct a model for $P_{true}(Y_{mis}|Y_{obs},R)$ for each unique $R$ in the dataset.
\item B) For each row of the dataset:
\begin{enumerate}
\item Generate samples from the model of $P_{true}(Y_{mis}|Y_{obs},R)$.
\item Generate samples from the conditional $\hat{P_k}(Y_{mis}|Y_{obs},R)$ estimated by the $k^{th}$ imputation algorithm for the current row.
\item Compute a score $s_k$ for the $k^{th}$ algorithm according to a chosen evaluation metric that compares the two conditional distributions using their corresponding samples.
\item Sort the scores to rank the $K$ algorithms for the current row.
\end{enumerate}
\item C) Use the method recommended by Demsar \cite{dem06} to combine the rankings for all the rows into a single average ranking over the $K$ algorithms. Confidence intervals are assigned to the average ranks to indicate which algorithms have significantly different imputation quality.
\end{itemize}

There are several choices possible for 1) the model of $P_{true}(Y_{mis}|Y_{obs},R)$, and 2) the metric for distribution comparison. Our evaluation framework is general in that it is not tied to any specific choice for either of these. Nevertheless in section~\ref{sec:evaluation} we discuss a few specific choices that are tractable and work well for high-dimensional data, e.g., problems with hundreds of features. For modeling $P_{true}(Y_{mis}|Y_{obs},R)$ we propose a Markov Random Field-based approach. We consider two choices: (a) a separate model for each missingness pattern; and (b) a single model for all missingness patterns. For distribution comparison metrics we consider 1) Kullback-Leibler (KL) divergence, 2) Symmetric KL divergence, and 3) two-sample testing for equality of two distributions. The challenge is that only samples of the true and imputed distributions are available. Recent advances in kernel-based approaches to estimating these metrics from samples have now made it possible to apply them to high-dimensional data. We estimate KL divergence using the convex risk minimization approach of \cite{ngu10}, symmetric KL using the cost-sensitive binary classification approach of \cite{rei11}, and perform two-sample testing using Maximum Mean Discrepancy (MMD) \cite{zar13,gre12,gre13}. In addition, we have proposed a new metric called Neighborhood-based Dissimilarity Score (NDS) that is faster to compute while providing similar results. We demonstrate the framework on several real datasets to compare four representative imputation algorithms from the literature.

\noindent\textbf{Contributions:} To our knowledge this is the first work that proposes the following.
\begin{itemize}
\item A principled framework for computing quantitative metrics to evaluate imputations and rank algorithms with statistical significance tests. Its generality allows different choices for distribution modeling and distribution comparison metrics to be plugged in according to the problem.
\item An efficient algorithm for jointly building distribution models for each unique missingness pattern.
\item Neighborhood-based Dissimilarity Score, which is two orders of magnitude faster than KL, symmetric KL and MMD, while giving similar results.
\end{itemize}
We can use this framework to evaluate the quality of imputation algorithms on different datasets, or understand the sensitivity of various imputation algorithms to different missingness types. The experiments presented in this paper are a first step towards such deeper investigations.

\noindent\textbf{Notes:} 1) The evaluation does not require knowledge of the missingness type, such as whether values are Missing Completely at Random (MCAR) or Not Missing at Random (NMAR) \cite{lit86}. It simply checks whether the samples generated by the imputation algorithm are distributed correctly compared to the true distribution. The imputation algorithm needs to be aware of the missingness type, but not the evaluation. 2) In our experiments we only consider up to 50\% missingness. Imputation is typically used for problems where the missingness percentage is not very high. At much higher percentages, e.g. 99\% missingness in Collaborative Filtering datasets, more specialized approaches are used and evaluation is focused more on predicting a few relevant missing entries rather than \emph{all} of them.

\noindent\textbf{Outline:} Section 2 presents two motivating examples that demonstrate our evaluation framework on datasets where it is possible to visually check the quality of the imputations. Section 3 explains the details of constructing the distribution models, the metrics used for comparing distributions, and statistical significance testing for the ranking of the algorithms. Section 4 presents the experimental results on several datasets, followed by a closing discussion in section 5.

\section{Illustrative Examples}
\label{sec:examples}
\begin{figure}[b!]
\centering
\includegraphics[width=0.9\linewidth]{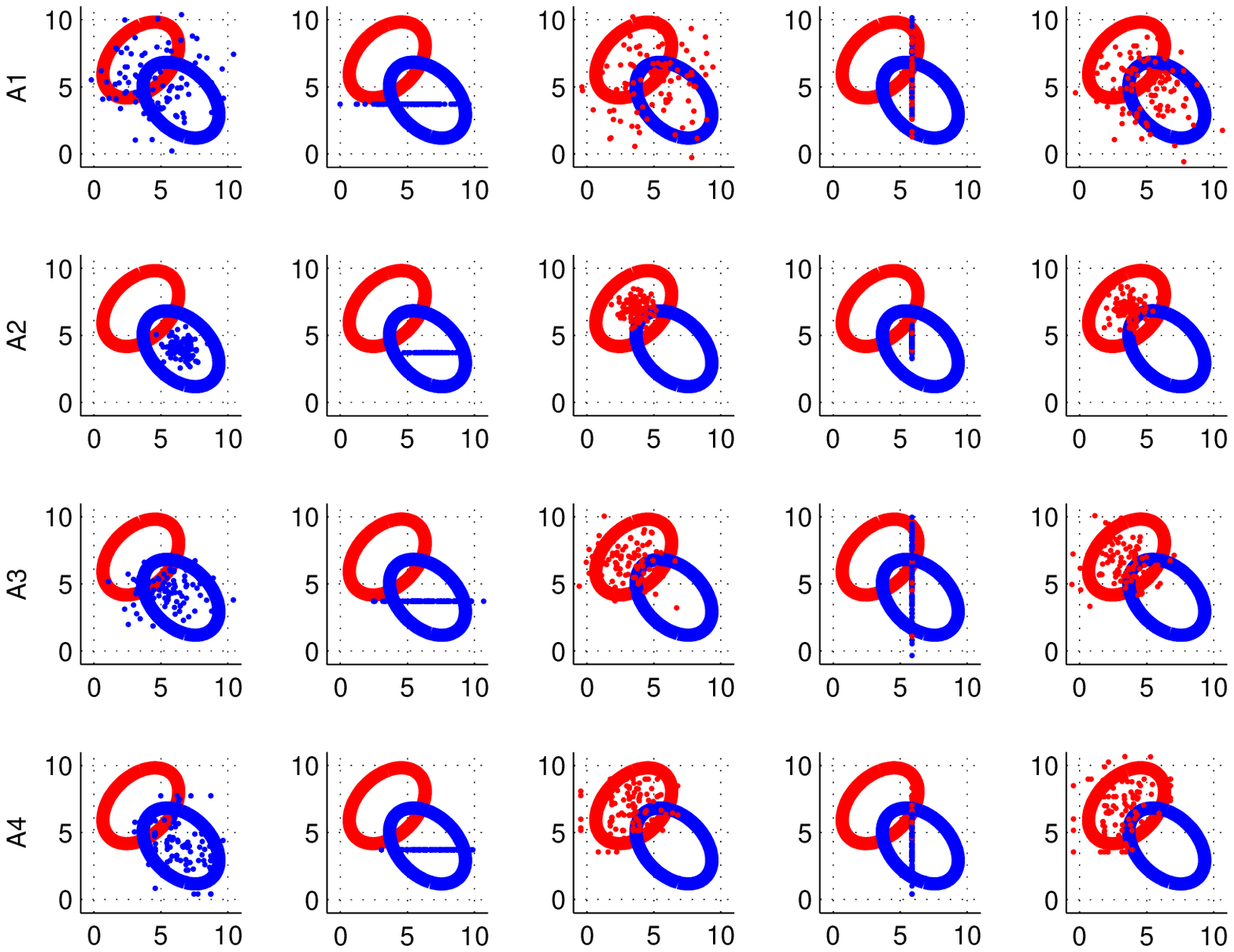}\\
\small{(a)}\\
\includegraphics[width=0.9\linewidth]{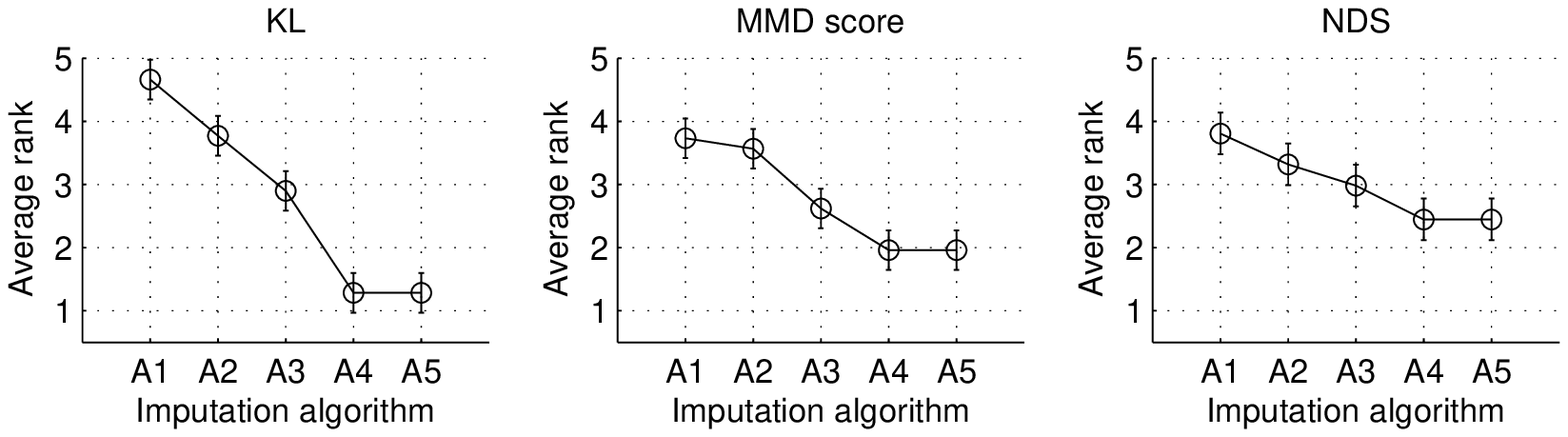}\\
\small{(b)}\\
\includegraphics[width=0.9\linewidth]{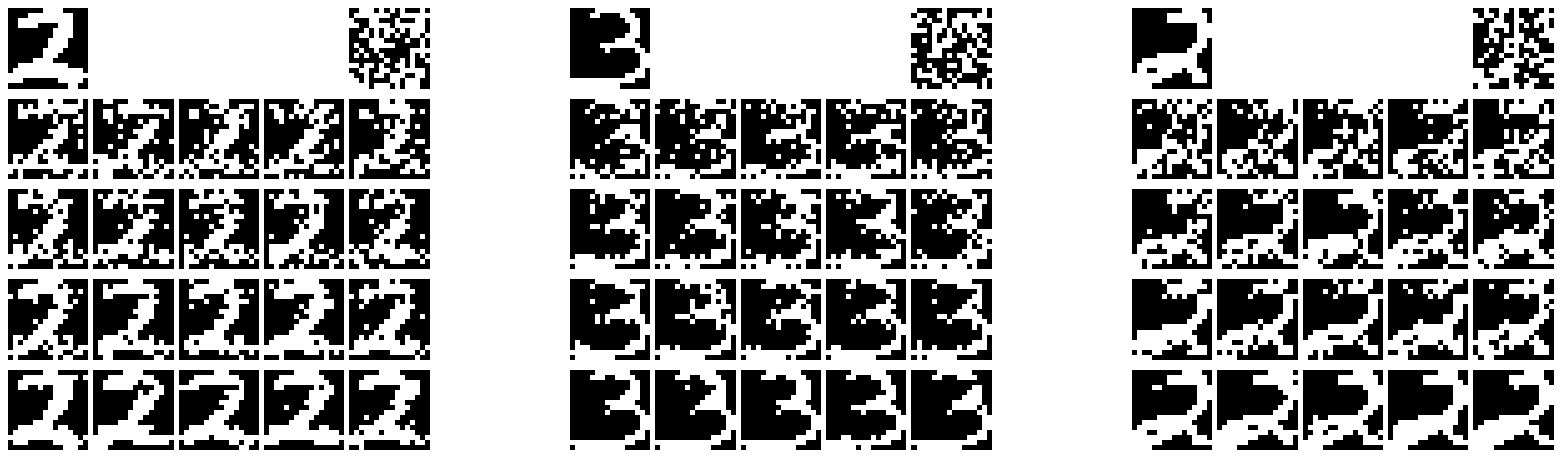}\\
\small{(c)}\\
\includegraphics[width=0.9\linewidth]{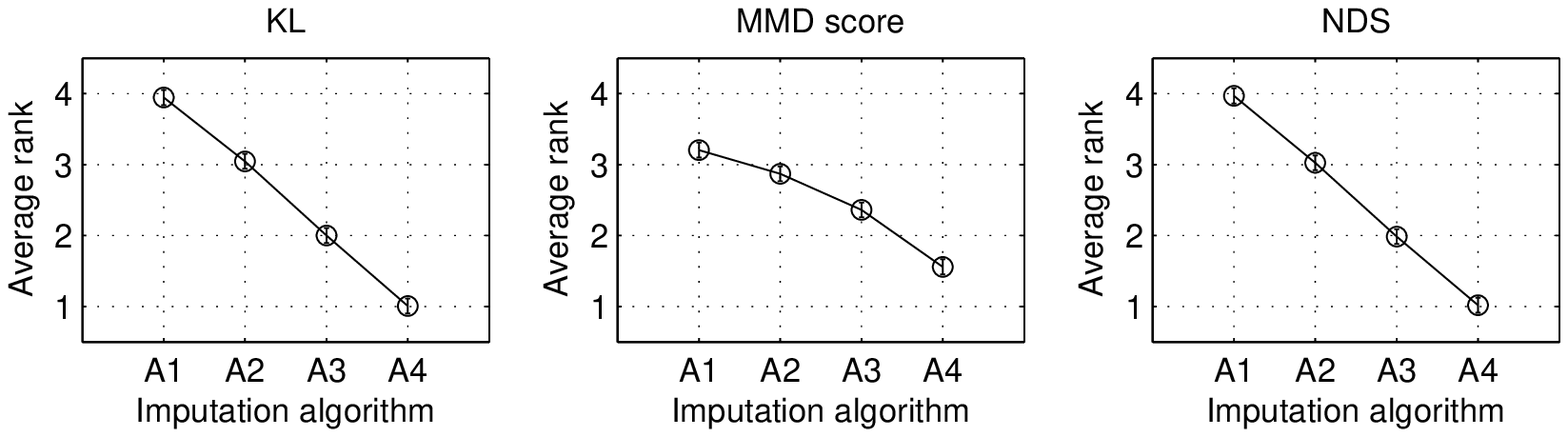}\\
\small{(d)}
\vspace{-0.1in}
\caption{\small Results of evaluating imputation algorithms on the Mixture of Gaussians (MoG) dataset and USPS. (a) Mixture components are shown as one-standard-deviation ellipses. Columns are different missing examples, rows are algorithms. The true conditional distribution of each column are: 1) $P(x,y|$\emph{label}=\emph{blue}$)$, 2) $P(x|y=4,$\emph{label}=\emph{blue}$)$, 3) $P(x,y|$\emph{label}=\emph{red}$)$, 4) $P(y|x=6,$\emph{label}=\emph{blue}$)$, 5) $P(x,y|$\emph{label}=\emph{red}$)$. The quality of the imputations can be judged by visually comparing the samples to the correct conditional distribution. (b) Average ranks computed by KL, MMD score, and NDS, for the full MoG dataset. (c) Each of three sets of images is a different USPS digit with missing values. The top row of each set shows the true image (left) and missing pixel indicators as a binary mask (right). The next four rows show five samples generated by A1 to A4 from top to bottom. (d) Average ranks computed by KL, MMD score, and NDS, for the full USPS dataset.}
\label{fig:usps_mog2d}
\end{figure}

Before explaining the details of our framework, we first show results on two datasets for which the imputations can be visualized. This allows a visual comparison of how good the different imputation algorithms are, which can then be checked against the evaluation by our framework.

\noindent\textbf{Datasets:} The first dataset is generated from a two-dimensional mixture of Gaussians with two components having equal prior probability. The Gaussians are shown in figure~\ref{fig:usps_mog2d} by their one-standard-deviation ellipses. The value of the component variable is also included in the dataset as a class label, color-coded in the figure by red and blue. We use 2000 data points sampled from the distribution. The second dataset is the USPS handwritten digits which contains $16 \times 16$ binary images flattened as 256-dimensional vectors. We select images for digits 2 and 3 to construct a dataset with 1979 rows and 256 columns. 50\% of the values are removed from both datasets completely at random. 

\noindent\textbf{Imputation algorithms \& Metrics:} The missing values are imputed using four different algorithms: Mean/Mode, Mixture Model, $k$ Nearest Neighbors ($k$NN), and MICE (labeled as $A1$, $A2$, $A3$, $A4$, respectively, in fig.~\ref{fig:usps_mog2d}). These algorithms are described in section~\ref{sec:evaluation}. For the 2D mixture of Gaussians, we have also included the true distribution itself as an imputation algorithm (labeled $A5$ in the plots) by using its conditionals to sample the missing variables given the observed ones. This is a good sanity check since a good metric should rank it as the best algorithm. We use KL divergence, MMD score, and NDS as the metrics (defined in section~\ref{sec:evaluation}).

\noindent\textbf{Results:} Figure~\ref{fig:usps_mog2d} summarizes the results for both datasets. To give a visual feel for the quality of the imputations, we have shown as examples a few specific data vectors and the samples generated by the algorithms to fill in the vectors. To interpret these plots, figure~\ref{fig:usps_mog2d} caption lists the variables that are missing for each example from the MoG dataset and explains which pixels are missing in the USPS images. The rank plots show the average rank attained by each algorithm over all the rows with missing values in the dataset. Error bars show 95\% confidence intervals for significant differences.

Visually judging the imputations, $A3$ ($k$NN) and $A4$ (MICE) produce the most plausible imputations for both datasets, while $A1$ (Mode/Mean) and $A2$ (Mixture Model) produce the worst ones. The ordering computed by all the three metrics in figure~\ref{fig:usps_mog2d} agree with this judgement. The true distribution for the MoG dataset has the best average rank for all three metrics, which is reassuring. Interestingly, MICE ties with the true distribution as the best algorithm. In the case of USPS, the ranking A1 to A4 (worst to best) is visually clear and is consistent with the average ranks and the small error bars. The results in section~\ref{sec:experiments} show a similar ordering of algorithms on datasets where the rankings cannot be verified in some other way. Doing this visual verification and corroborating its results with that of other datasets gives us more confidence that the evaluation framework is sensible.


\section{Evaluation Approach}
\label{sec:evaluation}
\vspace{-0.1in}
We provide details of the three key parts given in algorithm ${\mathcal A}$: (1) Model construction and sample generation (steps A and B1) (section 3.1), (2) Score computation using metrics (step B3) (section 3.2) and (3) Ranking of algorithms with statistical significance testing (step B4 and C) (section 3.3). We make the following assumptions. (a) There is a sufficient number of examples with no missing values for learning a good true model; this is true in many practical scenarios. (b) A conditional probabilistic model powerful enough to capture the underlying distribution can be built using these examples. (c) Generating samples from the model distribution can be done easily via Gibbs sampling. \\
\noindent{\bf Notations} Let ${\mathcal D}_N$ and ${\mathcal D}_M$ denote the sets of examples with no missing values and with missing values respectively. Let ${m}_j \subset \{1,\cdots,d\}$, $j=1,\cdots,J$ denote a subset of variables for which values are missing in ${\mathcal D}_M$ with $d$ denoting the number of variables. Note that each subset $m_j$ refers to a unique missing pattern and there are $J$ missing patterns in ${\mathcal D}_M$. Also, let $o_j = \{1,\cdots,d\} \setminus m_j$ be the observed variables. Let $Y^{(i)}_{m}$ and $Y^{(i)}_{o}$ denote the sets of missing and observed variables in the $i^{th}$ example respectively.

\subsection{Distribution Model Learning}
\vspace{-0.1in}
The first step is to build a distribution model such that for each missing pattern $m_j$, we can generate samples given the values for the corresponding observed pattern $o_j$. One possibility is to build a joint model $P(Y;\theta)$ using the set ${\mathcal D}_N$ (where $\theta$ is the model parameter). However, this may be hard and intractable; more importantly, it is not necessary. This is because we always have some observed variables that can be used to predict the missing variables. Furthermore, conditional models are relatively easier to build. Due to these reasons, for each $j$ we build a conditional distribution model $P(Y_{m_j}|Y_{o_j};\theta_j)$ for the missing pattern $m_j$ in ${\mathcal D}_M$ using the fully observed data set ${\mathcal D}_N$; here, $\theta_j$ denotes the $j^{th}$ model's parameter vector. \\
\noindent{\bf Model and Learning} While any good conditional model is sufficient for our method to work, we use the pairwise Markov random field (MRF) in this work. For ease of notation, let us remove the subscript from $m_j$. We use the model:
$P(Y_m|Y_o;\theta) = \frac{1}{Z(Y_o)} \exp(s(Y_m;\theta_m) + s(Y_{m};Y_{o},\theta_{mo}))$
where $Z(Y_o) = \sum_{Y_m} \exp(s(Y_m;\theta_m) + s(Y_{m};Y_{o},\theta_{mo}))$ is the partition function and, $s(Y_m;\theta_m)$ and $s(Y_{m};Y_{o},\theta_{mo})$ denote parametrized scoring functions involving self and cross features among the missing and observed variables. In our experiments, we use a simple linear model for the scoring function: $s(Y_m;\theta_m) = \sum_{k \in m} \sum_{l: {\bar y}_{l} \in {\mathcal Y}_k} \theta_{k,l} {\mathcal I}(y_{k} = {\bar y}_{l})$ where ${\mathcal Y}_k$ is the label space for the $k^{th}$ variable and ${\mathcal I}(\cdot)$ is the indicator feature function. The scoring function $s(Y_{m};Y_{o},\theta_{mo})$ is defined using similar feature functions involving a pair of variables and associated parameters. Given a missing pattern $m$, we construct a training set comprising of input-target pairs $\{(Y^{(i)}_o,Y^{(i)}_m):i=1,...,n\}$ from ${\mathcal D}_N$ and learn the model parameter $\theta$ using the piecewise pseudo-likelihood maximization training approach suggested by Sutton and McCallum (2008).

\noindent{\bf Samples from the distribution model:} Using the model parameters learned for each missing pattern $m_j$, we generate multiple imputations for each example in ${\mathcal D}_M$ (having the missing pattern $m_j$) via Gibbs sampling. For instance, sample generation for a discrete variable involves sampling a multinomial distribution with probabilities that can be easily computed from the conditional MRF distribution.

\noindent{\bf Single Model} One disadvantage of the above modeling is that we need to build $J$ conditional models. However, these models can be built in parallel and is a one time effort. Alternatively, we can build a single conditional model to deal with all missing patterns. The key difference is that we learn a single model parameter vector $\theta$ (instead of learning one model parameter vector $\theta_j$ per missing pattern $m_j$). In this case, there is only one dataset. The log likelihood term for each example involves different missing patterns. Therefore, the subset of model parameters used for each example is different, depending on the missing and non-missing variables involved; clearly, parameter overlap happens across the examples. Building a single model helps in reducing the computational complexity. Though such a model is expected to be inferior in prediction quality compared to the per missing pattern model, we found it to be adequate to meet the requirement of ranking the methods correctly.

\subsection{Metrics}
\vspace{-0.1in}
Given samples $(Y_m)$ from the model (\textit{true}) conditional distribution $(P)$ and multiple imputations $({\hat Y}_m)$ obtained from an imputation algorithm $k$ (with its distribution ${\hat P}_{k}$), we need a score/metric for measuring the the discrepancy between the samples $Y_m$ and ${\hat Y}_m)$. 
In this section we briefly describe the metrics (popular ones selected from the literature) used in our experiments. It is worth pointing out that the challenge here is the reliable estimation of metrics between two distributions using samples generated (independently) from them. Given its importance in applications, this problem has received much attention in the literature and several good methods have been proposed. Our work is the first one to make use of them for comparing missing value imputation algorithms.
For ease of notation in the discussion below, we drop the subscript $k$ indicating the dependency on the imputation algorithm, from ${\hat P}_k$. Below we assume that $|Y_m| = |{\hat Y}_m|)$. For all metrics other than KL divergence estimation, this condition can be relaxed.

\noindent{ \bf KL divergence via Convex Optimization:} 
Nguyen et al (2010) shows that the metric estimation problem can be posed as an empirical risk minimization problem and solved using standard convex programs. Estimating KL divergence is one instantiation of this problem; it is obtained using the solution of
the following optimization problem:
\begin{equation}
- min_{\alpha > 0} F = - \frac{1}{L} \sum_{u=1}^L log(L \alpha_u) + \frac{1}{2\lambda_L L} g(\alpha,Y_m,{\hat Y}_m)
\label{eq:KLD}
\end{equation}
where $\lambda_L$ is a regularization constant, $L$ is the number of imputation samples, $g(\alpha,Y_m,{\hat Y}_m) = \sum_{u,v} \alpha_u \alpha_v K(y^{(u)}_m,y^{(v)}_m) + \frac{1}{L} K({\hat y}^{(u)}_m,{\hat y}^{(v)}_m) - \sum_{u,v} \alpha_u \alpha_v K(y^{(u)}_m,{\hat y}^{(v)}_m)$ and $K(\cdot, \cdot)$ is a kernel function. We used the Gaussian kernel $K(y,{\hat y}) = \exp(-||y-{\hat y}||^2/\sigma)$ and set the kernel width parameter $\sigma$ to $1$ in our experiments. Finally, the KL divergence value is computed using the solution ${\hat \alpha}$ obtained from solving (\ref{eq:KLD}) as: $-\frac{1}{L}\sum_{u=1}^L log(L {\hat \alpha}_u)$.

\noindent{\bf Symmetric KL divergence via binary classification} It is known that KL divergence is not symmetric with respect to the distributions used, and, it is often useful to consider Jensen-Shannon divergence having the symmetric property (defined as: $\frac{1}{2}(KL(P,\frac{P+{\hat P}}{2}+KL(P,\frac{P+{\hat P}}{2})$). While it is theoretically possible to adapt the estimation approach suggested by Nguyen et al (2012) to estimate the symmetric KL divergence, it does not result in a simpler convex program like (\ref{eq:KLD}).
So we take a different route.
Reid et al (2011) developed a unified framework to show that many machine learning problems can be viewed as binary experiments and estimation of various divergence measures can be done by solving a binary classification problem with suitably weighted loss functions for the two categories. In our context, 
the symmetric KL divergence can be estimated as follows: (a) assign class labels +1 and -1 for the samples from the model distribution and imputation algorithm respectively, and, (b) build a classifier using the standard logistic loss function. The intuition is that, more divergent the distributions are, easier is the classification task. Therefore, the error rate measures the discrepancy between the distributions.

\noindent{\bf Maximum Mean Discrepancy} Gretton et al (2012) proposed a framework in which a statistical test can be conducted to compare samples from two distributions. The main idea is to compute a statistic known as maximum mean discrepancy (MMD) and use an asymptotic distribution of this statistic to conduct a hypothesis test ${\mathcal H}_0: P = {\hat P}$ with a suitably defined threshold. The MMD statistic is defined as the supremum over the mean difference between some function score computed from the samples of the respective distributions. Gretton et al (2012) shows that functions from reproducing kernel Hilbert space (RKHS) are rich for computing this statistic from finite samples and derives conditions under which MMD can be used to distinguish the distributions. Given the MMD function class ${\mathcal F}$ and the unit ball in a RKHS ${\mathcal H}$ with kernel $K(y,{\hat y})$ (e.g., Gaussian), the statistic is defined as:
\begin{equation}
{\hat MMD}^2_u({\mathcal F},Y_m,{\hat Y}_m) = g_{yy}(Y_m) + g_{{\hat y}{\hat y}}({\hat Y}_m) - g_{y{\hat y}}(Y_m,{\hat Y}_m)
\label{eq:mmd}
\end{equation}
where $$g_{y{\hat y}}(Y_m,{\hat Y}_m) = \frac{1}{L(L-1)} \sum_{u=1}^L\sum_{v\ne u}^L K({y}^{(u)}_m,{\hat y}^{(v)}_m);$$ $g_{yy}(Y_m)$ and $g_{{\hat y}{\hat y}}({\hat Y}_m)$ are similarly defined. To reduce the computational complexity and produce low variance estimates in the finite sample issue associated with computing (\ref{eq:mmd}), Zaremba et al (2013) proposes a statistical test known as B-test. Based on this theoretical backing, we pick up two metrics: (a) the statistic MMD and (b) the B-test result, for our study on ranking imputation algorithms. The B-test gives $0$ or $1$ as the output (with $1$ indicating acceptance of ${\mathcal H}_0$); therefore, there will be ties. However, the rank aggregation algorithm that we describe in section~\ref{sec:RA} can handle this.


\noindent{\bf Neighborhood Dissimilarity Score} One disadvantage associated with the metrics described above is that they are computationally expensive since we need to solve the associated problems for each example in ${\mathcal D}_M$. This becomes an issue especially when $J$ (the number of missing patterns) is large. We propose a neighborhood dissimilarity score (NDS) that is cheaper to compute and is defined as:
\begin{equation}
nds(h,Y_m,{\hat Y}_m) = \frac{1}{L^2}\sum_{u=1}^L \sum_{v=1}^L w_{uv} h(y^{(u)}_m,{\hat y}^{(v)}_m)
\label{eq:nds}
\end{equation}
where $h(\cdot)$ is a distance score (e.g., Hamming loss) computed between a pair of examples on the imputed values and $w_{uv}$ is a weighting factor given by: $\lambda^{h(\cdot)} + (1-\lambda)^{(h_{max} - h(\cdot))}$; $\lambda$ is a scaling parameter; and, $h_{max}$ is the maximum Hamming distance possible. Note that $w_{uv}$ gives more weight to neighbors and $\lambda$ controls the rate at which the weight falls off in terms of distance. We used $\lambda=0.1$ in our experiments.  
The overall complexity is $O(JL^2)$, but $L$ is typically small (e.g., 25). NDS is closely related to (\ref{eq:mmd}), but with three key differences: (a) we consider only the cross terms, (b) there are no kernel parameters to learn in NDS, and (c) the weighting is not uniform ($w_{uv}$ compared to the uniform weighting $\frac{1}{L(L-1)}$ used in (\ref{eq:mmd})). Our experimental results show that NDS is a good metric in the sense of producing results similar to MMD and KL based metrics.
\subsection{Rank Aggregation}
\vspace{-0.1in}
Let us take one metric ${\mathcal M}$. For each algorithm $k \in {\mathcal A}$ (a set of algorithms to compare), we compute the score $s^{(i)}_{k,{\mathcal M}}$ for the $i^{th}$ example in ${\mathcal D}_M$ using the procedures presented in the previous subsection. Using these scores, we rank the algorithms for each $i$. Collecting the ranks over all $i$ results in a rank matrix ${\mathcal R}_{\mathcal M}$ with each row representing an example and the columns representing the ranks of the algorithms in ${\mathcal A}$. To compare the algorithms, we draw analogy to the problem of statistically comparing multiple classifiers over multiple datasets (Demsar, 2006). In particular, we treat multiple imputations for each example as one dataset and we have scores for $|{\mathcal D}_M|$ datasets for a given algorithm (and the chosen metric). Our goal is to statistically compare the performance of algorithms in ${\mathcal A}$ on these datasets. Following Demsar (2006), we conduct the Friedman test with the null hypothesis that all algorithms are equivalent. This test involves computing the average ranks of each algorithm, followed by computing a statistic and checking against the ${\mathcal X}^2_F$ distribution for a given confidence level $\beta$ (we used 0.05). It turned out that the null hypothesis is always rejected for the algorithms that we considered. So, we conducted the Nemenyi test as the post-hoc test (Demsar, 2006). In this test, we compute the critical difference (CD) score, defined as: $q_{\beta} \sqrt{\frac{|{\mathcal A}|(|{\mathcal A}|+1)}{6|{\mathcal D}_M|}}$ where $q_{\beta}$ is obtained from a look-up table. The observed average rank difference between any two methods is statistically significant when it is greater than CD. We demonstrate this in our experiments later and make key observations.

\label{sec:RA}

\section{Experiments}
\label{sec:experiments}
We apply our framework to four UCI datasets and four representative imputation algorithms from the literature. We present the ranking results for the per-(missing)pattern model and show that the rankings are similar to those in section~\ref{sec:examples}. We then compare the results between the per-pattern model approach and the single model approach and show that the latter gives nearly identical results but is much cheaper to train. Then we consider various properties of the metrics, such as the self-consistency of the rankings for different choices of the ``true'' distribution, and their ability to discriminate among the algorithms with statistically significant rank differences.

\noindent\textbf{Datasets:} See table~\ref{tab:datasets}. Note that in the case of Mushroom, we selected the first 10 columns of the original dataset to keep the running times low. None of the datasets contain missing values originally. Values are removed completely at random, i.e. the probability of a variable missing is independent of its value or the value of other observed variables. The missing percentage is varied from 10 to 50\% in steps of 10. We use the full datasets without any missing values to build the distribution models.
\begin{table}
\begin{center}
\begin{tabular}{ | c | c | c | }
\hline
{\bf Dataset} & {\bf \#Rows} & {\bf \#Cols}\\
\hline
Flare & 838 & 9 \\
\hline
Mushroom & 2027 & 10 \\
\hline
SPECT & 213 & 22 \\ 
\hline
Yeast & 1200 & 14 \\ 
\hline
\end{tabular}
\end{center}
\caption{Datasets used for experiments}
\label{tab:datasets}
\end{table}

\begin{figure*}
\centering
\includegraphics[width=0.8\linewidth]{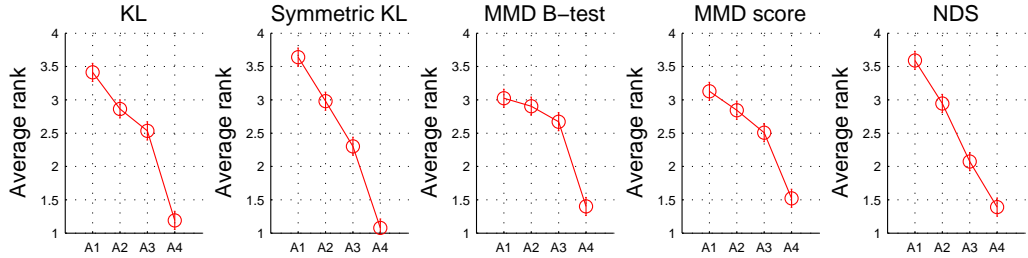}\\
\caption{Average rank of the four imputation algorithms on each of the five metrics for the Yeast dataset with 30\% entries missing at random}
\label{fig:per_row_results}
\end{figure*}

\begin{figure}
\centering
\includegraphics[width=0.9\linewidth]{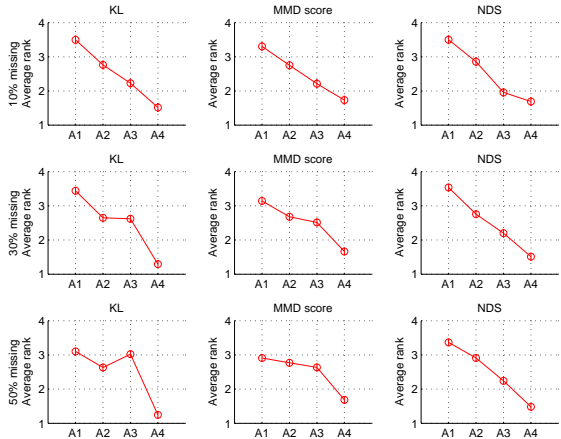}\\
\caption{Average rank of the four imputation algorithms on the KL (left column), MMD score (middle column) and NDS (right column) metrics, for the Yeast dataset with missing entries ranging from 10\% (top row), 30\% (middle row) and 50\% (bottom row).}
\label{fig:per_row_results2}
\end{figure}


\noindent\textbf{Imputation algorithms:} The following algorithms are included in the experiments:\\
\textbf{A1: Mode/Mean:} A simple baseline that samples from the marginal distribution of each column, fit by a Gaussian for continuous variables and a multinomial distribution for categorical variables.\\
\noindent\textbf{A2: Mixture Model:} A mixture model in which each mixture component is a product of univariate distributions. Inference and learning are done using Variational Bayes inference \cite{win05}.\\
\noindent\textbf{A3: $k$-Nearest Neighbors:} We implement the $k$-NN algorithm in \cite{has99}. Multiple samples are drawn from the distribution of values for the missing variable given by the set of nearest neighbors.\\
\noindent\textbf{A4: MICE:} Multiple Imputation using Chained Equations (MICE) \cite{buu11} is a state-of-the-art imputation algorithm which builds a predictor for each variable using all other variables as inputs. The learning alternates between predicting the missing values using the current predictors and updating the predictors using the current estimates of the missing values.\\
These algorithms are selected as representatives of commonly used approaches, rather than as a comprehensive set.

\noindent\textbf{Ranking results:} We have computed results using the per-pattern distribution models for 4 imputation algorithms, 5 metrics, 4 datasets, and 5 missing percentages, for a total of 400 combinations. The results can be analyzed along any of these dimensions to derive insight. For example, figure~\ref{fig:per_row_results} shows the results for the four different algorithms and five metrics for Yeast with 30\% missing values. All metrics pick MICE as the best algorithm. Or if we want to understand the effect of missing percentage on the performance of the imputation algorithms, we can compute the average ranks over all the datasets for each missing percentage and metric. Figure~\ref{fig:per_row_results2} shows the results for three missing percentages and three metrics. (We omitted some plots to save space). As the missing percentage increases, the algorithms become harder to distinguish under the KL and MMD score metrics, while NDS maintains discriminability. If we want to know the overall ranking of the imputation algorithms for each metric, we can aggregate the results across all datasets and all missing percentages. We observe that MICE is overall the best imputation algorithm, Mode/Mean is the worst, with Mixture Model and $k$-NN in between. These are some examples of the kind of analysis that become possible with our framework.

\noindent\textbf{Per-pattern vs. single models:} In section~\ref{sec:evaluation} we propose an alternative to learning a separate distribution model for each unique missingness pattern by learning all of them jointly with a single set of parameters. Here we verify whether learning a combined model in this manner introduces any approximation error that causes the average rankings to change. We re-run the same 400 combinations of algorithms, metrics, datasets, and missing percentages from the previous experiment, but now using the single distribution model instead of the per-pattern model. The percent difference in the average rank values between the two models for each metric is given in table~\ref{tab:per_pattern_vs_single}. The differences in the average rank values are too small to alter the rankings. (Note that the critical difference values do not depend on the choice of per-pattern vs. single model.) So both models give identical rankings over all metrics, datasets, and missing percentages.
\begin{table}
\begin{center}
\begin{tabular}{ | c | c | c | c | c |  }
\hline
{\bf KL} & {\bf Sym. KL} & {\bf MMD} & {\bf MMD} & {\bf NDS}\\
 &  & {\bf B-test} & {\bf score} & \\
\hline
2.27\% & 1.10\% & 1.40\% & 2.06\% & 0.36\%\\
\hline
\end{tabular}
\end{center}
\caption{\% change in average rank between per-pattern model and single model, averaged over all imputation algorithms, datasets, and missing percentages.}
\label{tab:per_pattern_vs_single}
\end{table}



\begin{figure}
\vspace{-0.1in}
\centering
\includegraphics[width=0.7\linewidth]{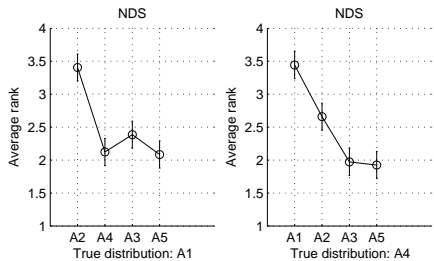}\\
\vspace{-0.1in}
\caption{Rankings of imputation algorithms with respect to different `true' distributions for Flare with 10\% missing values and NDS metric. The algorithm labels are: A1 = Mode/Mean, A2 = Mixture Model, A3 = $k$-NN, A4=MICE, A5 = Single model distribution.}
\label{fig:consistent1}
\end{figure}

\begin{figure}
\vspace{-0.15in}
\centering
\includegraphics[width=0.7\linewidth]{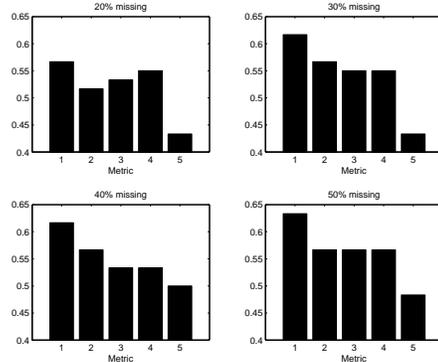}\\
\vspace{-0.2in}
\caption{Inconsistency scores for various metrics on Flare with different missing percentages. The metric labels are: MMD B-test = 1, MMD score = 2, SymmKL = 3, KL = 4, NDS = 5.}
\vspace{-0.2in}
\label{fig:consistent2}
\end{figure}

\noindent\textbf{Measuring metric inconsistency:} We can use our framework to investigate whether an evaluation metric has certain desirable properties, and based on the results decide on its usefulness. One such property we examine here is how ``inconsistent'' a metric is. 

Suppose we have $K$ algorithms $A_1$ to $A_K$. We can rank all algorithms except $A_i$ by treating its samples as coming from the `true' distribution and comparing the rest against it. $K$ such rankings can be computed for $i = 1$ to $K$. It is possible to check how inconsistent these rankings are with each other. Consider the example in figure~\ref{fig:consistent1}. When A1 is used as the true distribution, we see that A4 has a small average rank difference with A3 and A5, and a large difference with A2. When A4 itself is used as the true distribution, the ranking of \{A2, A3, A5\} follows the same pattern. A3 and A5 are ranked as closest to A4 while A2 is ranked as distant from A4. So for the pair (A1,A4) the ranking of \{A2, A3, A5\} given by the average rank differences in the left plot matches the ranking in the right plot. When they do not match, we call them \emph{inconsistent}.

We can quantify the inconsistency using the Kendall-Tau distance between the two rankings for every pair of algorithms and averaging the distances. Figure~\ref{fig:consistent2} shows the inconsistency scores computed for five metrics on Flare for different missing percentages. Note that NDS has the lowest inconsistency score among all the metrics. The difference between NDS and other metrics is particularly large at the lower missing percentages. Similar results are observed on other datasets also. This provides further evidence that NDS is a good metric for evaluating imputations.

\section{Conclusion}
We presented a framework to quantitatively evaluate and rank the goodness of imputation algorithms. It allows users to (a) work with any decent conditional modeling choice, (b) use metrics that are efficient and consistent (any improved quality metrics that appear in future as this field matures can be easily incorporated), (c) compare metrics using any new consistency checks, and (d) rank algorithms using well-laid out statistical significance tests. 


\small
\bibliography{paper}
\bibliographystyle{icml2014}

\end{document}